\documentclass{midl} 

\usepackage{lineno}
\usepackage{booktabs}
\usepackage{natbib}
\usepackage{multirow}
\usepackage{caption}
\usepackage{mwe}
\usepackage{rotating}
\usepackage{float}                     
\floatstyle{ruled}\restylefloat{algorithm}
\usepackage{algorithmicx}
\usepackage{algpseudocode}
\usepackage{xcolor}
\usepackage[T1]{fontenc}

\usepackage{algpseudocode,algorithm}
\jmlrvolume{}
\jmlryear{2026}
\jmlrworkshop{MIDL 2026}

\title[Diffusion-Driven Fine-Grained Nodule Synthesis]
{A Diffusion-Driven Fine-Grained Nodule Synthesis Framework for Enhanced Lung Nodule Detection from Chest Radiographs}

\midlauthor{%
  \Name{Aryan Goyal\midljointauthortext{Contributed equally}\nametag{$^{1}$}} \Email{garyan18@gmail.com}\\%
  \Name{Shreshtha Singh\midlotherjointauthor\nametag{$^{1,2}$}} \Email{shreshtha.singh@qure.ai}\\%
  \Name{Ashish Mittal\midlotherjointauthor\nametag{$^{1}$}} \Email{ashish.mittal@qure.ai}\\%
  \Name{Manoj Tadepalli\nametag{$^{1}$}} \Email{manoj.tadepalli@qure.ai}\\%
  \Name{Piyush Kumar\nametag{$^{1}$}} \Email{piyush.kumar@qure.ai}\\%
  \Name{Preetham Putha\nametag{$^{1}$}} \Email{preetham.putha@qure.ai} \\
\addr $^{1}$ Qure.ai, India \\%
  \addr $^{2}$ Indian Institute of Technology, Bombay%
}

\begin{document}

\maketitle


\begin{abstract}
Early detection of lung cancer in chest radiographs (CXRs) is crucial for improving patient outcomes, yet nodule detection remains challenging due to their subtle appearance and variability in radiological characteristics like size, texture, and boundary. For robust analysis, this diversity must be well represented in training datasets for deep learning based Computer-Assisted Diagnosis (CAD) systems. However, assembling such datasets is costly and often impractical, motivating the need for realistic synthetic data generation. Existing methods lack fine-grained control over synthetic nodule generation, limiting their utility in addressing data scarcity. This paper proposes a novel diffusion-based framework with low-rank adaptation (LoRA) adapters for characteristic controlled nodule synthesis on CXRs. We begin by addressing size and shape control through nodule mask conditioned training of the base diffusion model. To achieve individual characteristic control, we train separate LoRA modules, each dedicated to a specific radiological feature. However, since nodules rarely exhibit isolated characteristics, effective multi-characteristic control requires a balanced integration of features. We address this by leveraging the dynamic composability of LoRAs and revisiting existing merging strategies. Building on this, we identify two key issues, overlapping attention regions and non-orthogonal parameter spaces. To overcome these limitations, we introduce a novel orthogonality loss term during LoRA composition training.
Extensive experiments on both in-house and public datasets demonstrate improved downstream nodule detection. 
Radiologist evaluations confirm the fine-grained controllability of our generated nodules, and across multiple quantitative metrics, our method surpasses existing nodule generation approaches for CXRs. \\ 
\textbf{Keywords}: Lung Nodule Synthesis, Chest Radiograph, Diffusion Models , LoRA , LoRA Merging
\end{abstract}

\begin{figure}[t]
\centering
\setlength{\tabcolsep}{1pt} 
\renewcommand{\arraystretch}{2} 
\begin{tabular}{ccccc}
\scriptsize \textbf{Ground Truth} & \scriptsize \textbf{ACGAN} & 
\scriptsize \textbf{ReACGAN} & 
\scriptsize \textbf{CR-Fill} & 
\scriptsize \textbf{Ours} \\[-3pt]
\includegraphics[width=0.19\linewidth]{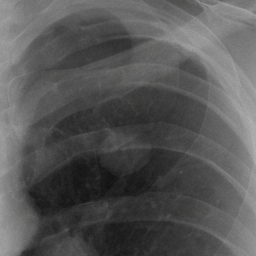} &
\includegraphics[width=0.19\linewidth]{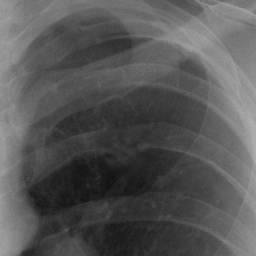} &
\includegraphics[width=0.19\linewidth]{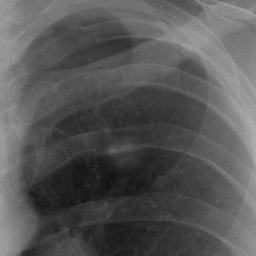} &
\includegraphics[width=0.19\linewidth]{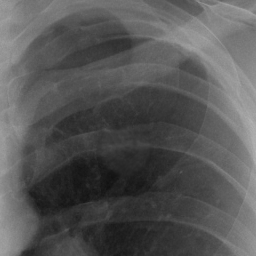} &
\includegraphics[width=0.19\linewidth]{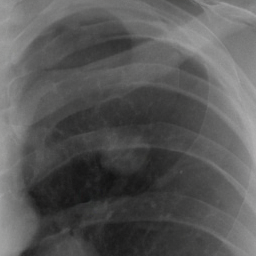} \\
\end{tabular}
\vspace{-10pt}
\caption{Comparison of nodule generations across different methods: ACGAN\cite{odena2017acgan}, ReACGAN\cite{lee2021reacgan}, CR-Fill\cite{zhao2021crfill}, and Ours.}
\label{fig:gan_comparison}
\end{figure}
\section{Introduction}
\label{sec:intro}
Lung cancer remains a leading cause of cancer mortality \citep{lc_1}. 
Early detection is critical and  localized tumors can reach five-year survival 
rates above 70\%~\citep{seer_csr_1975_2014}. Despite advances in imaging, many cases are 
first suspected on chest X-rays (CXR), one of the most common tests 
\citep{Rogers2010}; accurate nodule detection on CXR is therefore essential. 
Pulmonary nodules exhibit key characteristics such as size, calcification, 
border definition, and homogeneity that are essential for malignancy assessment 
\citep{albert2009solitary}. Robust CAD requires datasets spanning these traits, 
yet high-quality annotated CXRs are scarce, manual labeling is labor-intensive, 
costly, and suffer from intra-observer variability \citep{lc_5}. Synthetic augmentation can expand training 
data and improve detectors \citep{schultheiss2021, hanaoka2024, wang2022synthetic}, 
but current nodule synthesis or inpainting approaches lack fine-grained control over nodule characteristics. Given the high CXR miss 
rate, precise and clinically faithful control of nodule attributes is needed.

\noindent Diffusion models offer strong potential for precise generative control, outperforming GANs in image quality, stability, and diversity \cite{dhariwal2021diffusion, yang2022survey}. Among them, Diffusion Transformers (DiT) \cite{peebles-2023} achieve superior performance compared to large UNet-based architectures such as SDXL \cite{podell2023sdxl}. Controllability in diffusion models has advanced through lightweight conditioning mechanisms, including ControlNet \cite{zhang2023controlnet} and LoRA-based methods \cite{hu2021lora, lora-diffusion-2023, lyu2023lowrank}, which steer generation toward specific attributes without full-model retraining. Concept Sliders \cite{gandikota-2023} further enable continuous and compositional control by learning semantic directions in latent activation space. 

\noindent In our work, we adopt DiT as our backbone and condition it with binary masks to control nodule shape, size, and location. After training the backbone, we attach LoRA modules \cite{hu2021lora} for four radiological attributes:  calcification, border definition (regular/irregular), homogeneity, and subtlety by using characteristic-specific subsets to capture fine-grained distinctions without affecting general nodule synthesis. Since nodules often exhibit multiple attributes, we explore LoRA combination strategies like LoRA-Switch \cite{kong2024loraswitchboostingefficiencydynamic}, linear merging \cite{prabhakar2024lorasoupsmergingloras}) and training-based fusion ZipLoRA \cite{shah2023ziplorasubjectstyleeffectively}. We find these methods limited by spatial competition and interference from non-orthogonal adapter weights. To address this, we propose a training-based merging strategy with a Frobenius norm penalty that encourages orthogonality across LoRA matrices.

\noindent The proposed framework (i) introduces a novel diffusion-based approach for generating synthetic lung nodules on CXRs, (ii) enables controllable synthesis of key radiological characteristics through characteristic-specific and merged LoRA adapters, and (iii) is validated through extensive experiments on both in-house and public datasets, demonstrating consistent improvements over existing methods, including downstream detection gains with AUCs of 0.90 on JSRT and 0.93 on CheXray14, further supported by radiologist evaluations. The code will be  released at:
\url{https://github.com/shreshthasingh00/Nodule-Crafter-Diffusion-driven-Nodule-synthesis-on-CXR/}

\section{Related Work}

\textbf{Lung Nodule Synthesis:}
Early work synthesized nodules by forward-projecting CT-derived annotations onto radiographs \citep{schultheiss2021, litjens-2010, behrendt2023systematic}. Other methods generate nodules directly in masked CXR regions using inpainting \citep{sogancioglu2018chestxrayinpaintingdeep} or feature-level blending \citep{gundel-2021}. GAN-based approaches \citep{shen-2022} enable factorized control over shape, size, and texture but lack fine-grained characteristic manipulation. To the best of our knowledge, no prior work has investigated diffusion models for fine-grained, controllable nodule synthesis in CXRs.

\noindent \textbf{Controllable Image Generation:}
Early controllability in diffusion models was achieved through classifier and classifier-free guidance \citep{dhariwal2021diffusion, ho2022classifierfreediffusionguidance}, followed by lightweight conditioning modules such as T2I-Adapters and ControlNet \citep{mou2023t2iadapter, zhang2023controlnet}. Editing and personalization methods, including prompt-to-prompt and DreamBooth \citep{hertz2022prompt, ruiz2023dreambooth}, enabled localized and concept-specific control.
Among these, LoRA \citep{hu2021lora} has emerged as a dominant mechanism for controllable generation as it enables efficient low-rank fine-tuning and allows for  composability of  adapter modules  

\noindent \textbf{LoRA Merging: } Merging multiple LoRA adapters remains challenging, with works such as ZipLoRA, Mix-of-Show, and K-LoRA \citep{shah2023ziplorasubjectstyleeffectively, gu2023mixofshowdecentralizedlowrankadaptation, ouyang2025kloraunlockingtrainingfreefusion} showing that naive fusion leads to concept conflicts, loss of identity, and attenuation of fine details. Several approaches aim to mitigate these issues: DO-Merging \citep{zheng2025decoupleorthogonalizedatafreeframework} enforces layer-wise orthogonalization of LoRA directions, LoRI \citep{zhang2025lorireducingcrosstaskinterference} reduces cross-task interference via sparse masking and frozen projections, and ZipLoRA further introduces trainable merger coefficients to balance layer-wise adapter contributions. These methods attempt to resolve conflicts after independent trainings. In contrast, our method integrates a Frobenius norm based orthogonality loss directly into the training of each characteristic-specific adapter, ensuring that the learned LoRAs are inherently compatible for merging. 

\section{Datasets}

\noindent \textbf{Nodule Characteristics Definitions:} 
Pulmonary nodules exhibit several radiological attributes important for distinguishing them from mimickers and assessing malignancy. Their size ranges from a few millimeters up to 3 cm. Calcification, arising from calcium deposits, is frequently associated with benignity \cite{lc_7}. Border definition reflects edge morphology: regular, well-defined margins are typically stable, whereas irregular, spiculated, or lobulated borders may indicate malignancy \cite{lc_8}. Homogeneity describes texture uniformity; homogeneous nodules show consistent intensity, while inhomogeneous ones exhibit variation due to necrosis or vascularity, features often linked to malignant processes \cite{Balagurunathan2019}. Perceptual subtlety is also critical, as nodules may be faint or obscured by ribs and vessels, making detection challenging.

\noindent \textbf{In-house and Public Datasets: }
Our in-house dataset comprises 1.2M frontal-view CXRs from partner hospitals, including 40k chest xrays with pulmonary nodules. Each nodule is delineated with shape annotations and labeled for calcification (7,875), regular border (10,424), irregular border (5,153), homogeneous texture (4,640), inhomogeneous texture (5,883), and subtlety (5,000 cases graded 1–5), with all annotations independently provided by three experienced radiologists. We further split the inhouse trainset for generation and downstream trainings.
For evaluation, we additionally use the public \textit{ChestX-ray14} \cite{kufel-2023} and \textit{JSRT} \citep{shiraishi1996computer} datasets, where JSRT also includes subtlety scores (1 = most subtle, 5 = most obvious) for nodules. Both public datasets provide nodule bounding boxes; we refine these using our segmentation model trained on the in-house data and align predictions with the provided boxes. All selected segmentations were reviewed by radiologists for consistency. Dataset statistics for all the datasets are provided in Table~\ref{tab:test-dataset}. 
\begin{table}[h]
  \centering
  \caption{Summary of datasets used for training and evaluation.}
  \label{tab:test-dataset}

  \scriptsize
  \setlength{\tabcolsep}{6pt}
  \renewcommand{\arraystretch}{0.9}

  \begin{tabular}{@{} l cccc cc @{}}
    \toprule
      & \multicolumn{3}{c}{\textbf{In-house}} 
      & \multicolumn{1}{c}{\textbf{JSRT}}
      & \multicolumn{1}{c}{\textbf{CheX-ray14}} \\
    \cmidrule(lr){2-4}

      & \textit{Diffusion Trainset} 
      & \textit{Downstream Trainset} 
      & \textit{Testset} 
      & \textit{Testset}
      & \textit{Testset} \\

    \midrule
    \textbf{Total samples}  
      & 1{,}100{,}000 
      & 100{,}000 
      & 12{,}000 
      & 247 
      & 500 \\

    \textbf{Nodule samples} 
      & 28{,}000 
      & 10{,}000  
      & 2{,}000  
      & 154 
      & 66 \\
    \bottomrule
  \end{tabular}
\end{table}

\begin{figure}[t]
\centering
\begin{tabular}{ccc}

{\scriptsize (a) Original patch } & {\scriptsize (b) Nodule mask} & {\scriptsize (c) Generated Patch} \\[2pt]
\includegraphics[width=0.15\linewidth]{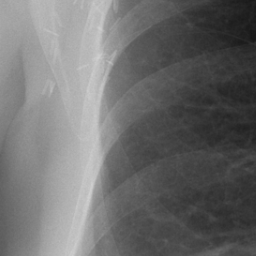} &
\includegraphics[width=0.15\linewidth]{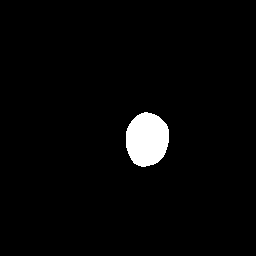} &
\includegraphics[width=0.15\linewidth]{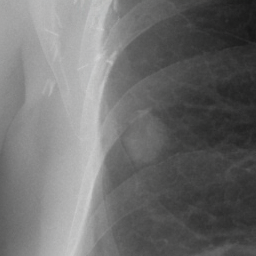} \\[-1pt]
\end{tabular}
\caption{ Diffusion Backbone generation on a chest X-ray patch:  given an original CXR patch and a binary nodule mask, the model generates a nodule within the masked region}
\label{fig:Diffusion baseline generation}
\end{figure}

\begin{figure*}[t]
    \centering
    \includegraphics[width=\textwidth]{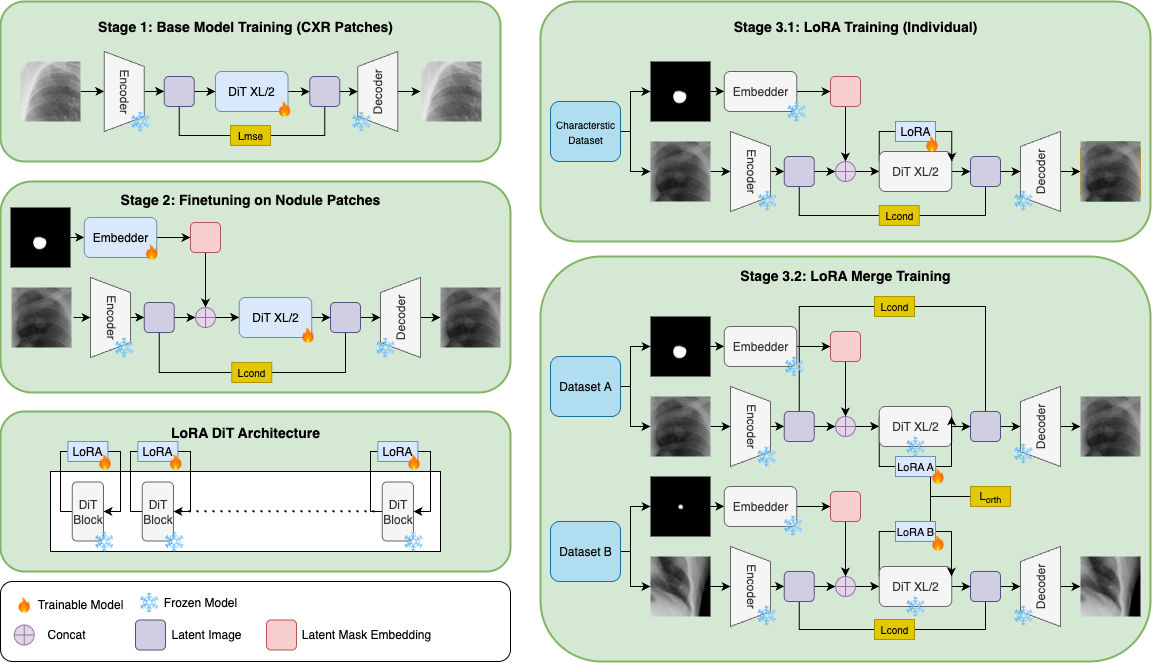}
    \caption{Overview of the pipeline with characteristic control. Training proceeds in 3 stages: \textbf{Stage 1 (Base Model Training)}-pre-train DiT-XL/2 on 2M of CXR patches; \textbf{Stage 2 (Nodule Patch Fine-tuning)}-finetune on nodule-centered patches with binary nodule masks for localized synthesis; \textbf{Stage 3.1 (Characteristic-Specific LoRA Training)}-train individual LoRA adapters (calcification, homogeneity, border regularity, subtlety) with the backbone frozen, using characteristic-curated datasets; \textbf{Stage 3.2 (LoRA Merge Training)}-jointly train selected adapters with an orthogonality regularizer}
    \label{fig:your_label}
\end{figure*}

\section{Methodology}
\subsection{Background}
\textbf{Diffusion Models} are a class of generative models that learn to approximate complex data distributions by iteratively transforming random noise into structured samples. They operate through a two-phase process: a \textit{forward diffusion process}, in which Gaussian noise is incrementally added to training data over a sequence of steps, and a \textit{reverse denoising process}, where a network is trained to recover the original data by gradually removing the noise. This learned denoising procedure allows the model to sample from the target distribution by reversing the noising trajectory. The forward diffusion process is modeled as a Markov chain, where Gaussian noise is added at each time step. Given an initial data sample \( x_0 \), the noised sample at time step \( t \) is computed as:\begin{equation}
q(x_t | x_{t-1}) = \mathcal{N}(x_t; \sqrt{1 - \beta_t} x_{t-1}, \beta_t I)\end{equation} where \( \beta_t \) is the noise schedule controlling the amount of noise added at each step. The final state \( x_T \) approaches a pure Gaussian noise distribution. 
To train the model, we optimize:



\begin{equation}
L_{\text{cond}} = \mathbb{E}_{x_0, t, \epsilon, c} \left[ \| \epsilon - \epsilon_\theta(x_t, t, c) \|^2 \right]
\end{equation}where \( \epsilon_\theta(x_t, t, c) \) represents the model’s prediction of the noise at time \( t \) and condition \( c \). \\

\begin{figure*}[t]
\centering
\setlength{\tabcolsep}{2pt} 
\renewcommand{\arraystretch}{0.7} 

\begin{tabular}{ccccc}

{\small (a) Regular Border } & {\small (b)Irregular Border} & {\small (c) Homogeneous} & {\small (d) Inhomogeneous} & {\small (e) Calcification} \\[1pt]
\includegraphics[width=0.18\textwidth]{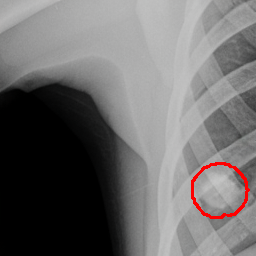} &
\includegraphics[width=0.18\textwidth]{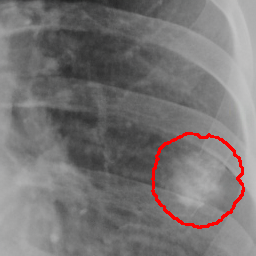} &
\includegraphics[width=0.18\textwidth]{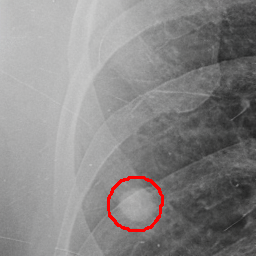} &
\includegraphics[width=0.18\textwidth]{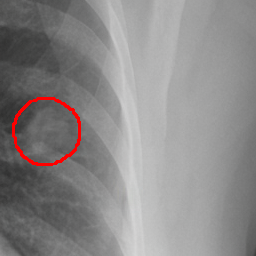} &
\includegraphics[width=0.18\textwidth]{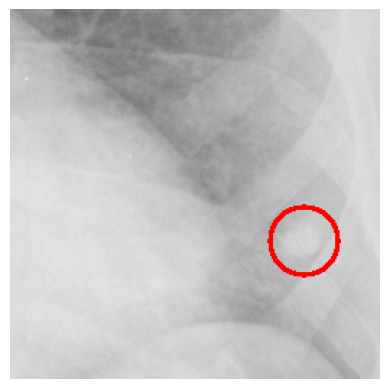} \\[2pt]

\includegraphics[width=0.18\textwidth]{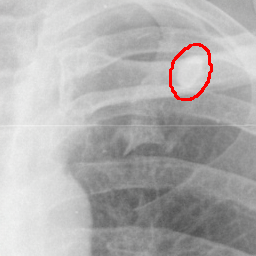} &
\includegraphics[width=0.18\textwidth]{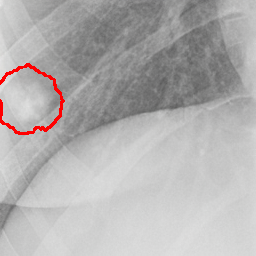} &
\includegraphics[width=0.18\textwidth]{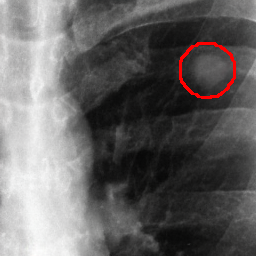} &
\includegraphics[width=0.18\textwidth]{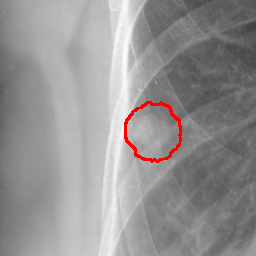} &
\includegraphics[width=0.18\textwidth]{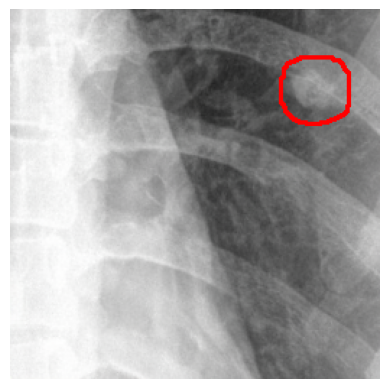} \\

\end{tabular}

\caption{
\small
Results of characteristic-specific LoRA training: (a) nodule with a regular border, (b) nodule with an irregular border, (c) nodule with homogeneous texture, (d) nodule with inhomogeneous texture, and (e) nodule with calcification.
}
\label{fig:nodulecharwiselora}
\end{figure*}

\subsection{ Diffusion Backbone}

We build our framework on a Diffusion Transformer leveraging its capacity for large scale data and high-fidelity generation. We first pre-train DiT-XL/2 on approximately 2 million unlabeled chest radiograph patches, enabling the model to capture the semantic structure of thoracic anatomy and generate realistic CXR patches. To specialize the model for localized nodule synthesis, we finetune it on curated nodule patches using binary masks as spatial conditioning signals. These masks define the target region for nodule placement, allowing the model at inference to synthesize nodules at desired locations through inpainting while preserving surrounding lung structures as shown in Figure \ref{fig:Diffusion baseline generation}. To further improve adherence to the conditioning masks, we apply classifier-free guidance(CFG) during inference. 

\noindent \textbf{Rationale for Separate LoRA Adapters:}
Using CFG to control multiple semantic labels simultaneously competes with mask-based conditioning and degrades boundary fidelity. Empirically, applying CFG to both masks and labels produced suboptimal results. Consequently, we adopt LoRA modules for characteristic control as they are computationally inexpensive while using CFG for mask conditioning. A comparison between CFG based label control and our separate LoRA approach is provided in Appendix~\ref{sec:cfgvslora}.


\subsection{Characteristic-Specific LoRA Adapters}
\begin{figure}[t]
\centering
\begin{tabular}{ccc}
{\scriptsize (a) Subtlety ($\alpha$=24)} & {\scriptsize (b) Subtlety ($\alpha$=16)} & {\scriptsize (c) Subtlety ($\alpha$=12)} \\[4pt]
\includegraphics[width=0.15\linewidth]{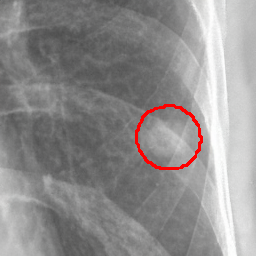} &
\includegraphics[width=0.15\linewidth]{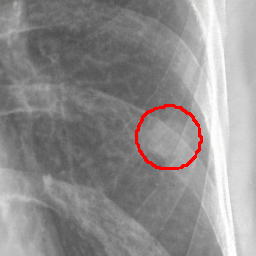} &
\includegraphics[width=0.15\linewidth]{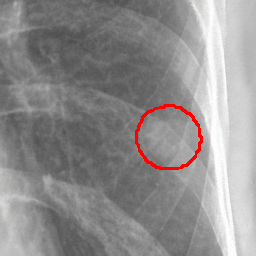} \\[-2pt]
\end{tabular}
\caption{Subtlety slider generations on a CXR patch. From left to right: decreasing LoRA scale $\alpha$ values produce more subtle nodules.}
\label{fig:scale_comparison}
\end{figure}

We design characteristic-specific LoRA adapters for clinically relevant nodule attributes.  LoRA freezes the pre-trained weights and learns a compact set of rank-decomposed updates, significantly reducing the number of trainable parameters.  Given a pre-trained weight matrix \( W_0 \), LoRA parameterizes its update as \( \Delta W = A B \), where \( A \in \mathbb{R}^{d \times r} \) and \( B \in \mathbb{R}^{r \times k} \), with \( r \ll \min(d, k) \).  Each characteristic LoRA adapter is trained on samples curated for its respective characteristic, while the DiT-XL/2 backbone remains frozen. Qualitative results for individual characteristics are presented in Figure~\ref{fig:nodulecharwiselora}. For the subtlety attribute, we extend the Concept Sliders~\cite{gandikota-2023} approach to leverage the graded annotations in our dataset. Radiologists labeled each nodule on a discrete scale from 1 (most subtle) to 5 (most obvious). To incorporate this into training, we map the annotation level directly to the LoRA scale parameter ($\alpha$). Nodules with lower subtlety scores are assigned smaller $\alpha$ values, yielding weaker updates and less visible nodules, whereas higher scores correspond to larger $\alpha$ values, amplifying the updates and making nodules more apparent.
\begin{equation}
    W = W_{0} + \alpha(s) \, \Delta W
\end{equation}

\noindent where $s \in \{1,2,3,4,5\}$ is the radiologist-assigned subtlety score, and we take $\alpha(s) = 2^{2+s}$ as the LoRA scale which we found to work well empirically during subtlety slider training. During inference, we set $\alpha$ for generating nodules at  different subtlety levels as shown in Figure \ref{fig:scale_comparison}.

\subsection{Challenges in Multi-Characteristic Synthesis}

Conventional merging strategies such as linear merge and switching~\cite{zhong2024multiloracompositionimagegeneration} assume that LoRA updates combine uniformly across layers. In practice, this leads to sub-optimal synthesis like one characteristic often dominates and artifacts often appear near mask boundaries. In general, LoRA weights are highly sparse (60--70\% near zero; $|w| < 0.01$), meaning a small subset of parameters drives perceptual change, making naive averaging brittle~\cite{ouyang2025kloraunlockingtrainingfreefusion}. Two factors contribute most: \textbf{overlapping attention regions} and \textbf{non-orthogonal updates}. Unlike natural-image settings where attributes may be spatially distinct, all adapters operate on the same nodule region, causing competing updates. 
 Our Frobenius-norm orthogonality analysis~\cite{Nakayama1952Orthogonality} further shows that independently trained adapters are not well separated, leading to correlated updates. This non-orthogonality causes interference regardless of the merging strategy employed.

\subsection{Orthogonality-Constrained Adapter Merging}
\begin{figure}[t]
    \centering
    \small
    \includegraphics[width=0.4\linewidth]{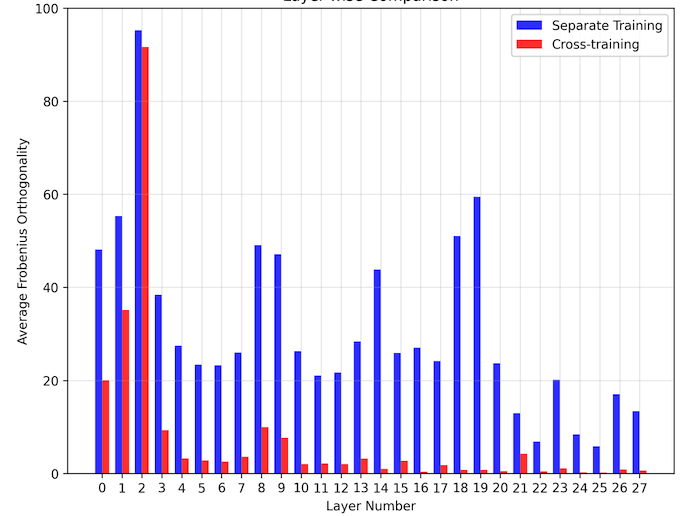}
    \caption{Layer-wise Frobenius norm comparison of cross-trained vs.\ separately trained adapters across two characteristic pairs.}
    \label{fig:orthogonality-layers}
\end{figure}

To reduce interference, we promote orthogonality between adapter subspaces during training. For adapters with updates  $W_1$ and $W_2$, we add \begin{equation} \mathcal{L}_{\text{orth}} = \| W_1 W_2^\top \|_F^2 \end{equation}
and apply the term pairwise when using more than two adapters. Although the formulation naturally extends to merging multiple characteristics, in this work, we evaluate the setting of two-attribute combinations.
This loss penalizes correlations between the parameter subspaces of different adapters. We scale this term in the final loss using coefficient $\lambda$, allowing a trade-off between orthogonality and reconstruction fidelity.  With this constraint, adapters compose reliably: linear averaging suffices, and each characteristic remains controllable via its scalar $\alpha$. As illustrated in Figure~\ref{fig:lora-merge-rotated}, our merging follows the nodule mask better preserving its structure and follows the characteristics better than just inference time linear merging. We also analyse the orthogonality across 28 trasformer layers shown in  Figure \ref{fig:orthogonality-layers}, which proves that with cross training the Frobenius norms are much lower compared to separately trained adapters.



\label{subsec:ortho}

\section{Results and Analysis}

We evaluate our framework along three criteria: 
(i) radiologist evaluation , 
(ii) quantitative assessment on standard image metrics, 
(iii) downstream impact on nodule detection models





\subsection{ Radiologist Evaluation } 
\noindent \textbf{Task 1 (Realism)} We presented a balanced set of 50 nodules (25 real and 25 generated by our base diffusion model) to three expert radiologists. On average, \textbf{90\%} of real nodules were correctly identified as real, while \textbf{80\%} of synthetic nodules were also labeled as real, indicating that our model produces nodules that are highly realistic and often indistinguishable from genuine cases.

\noindent \textbf{Task 2 (Controllability)} To evaluate characteristic-specific LoRA adapters, we generated 10 nodules per target feature (e.g., border type, texture) and asked 3 radiologists to verify the intended trait. Table~\ref{tab:characteristic-accuracy} summarizes the majority-agreement rates across features. 




\noindent \textbf{Task 3 (Subtlety)} We evaluated Subtlety LoRA by generating 20 synthetic nodule patches from the same mask, each rendered at 3 different levels, as shown in Figure \ref{fig:scale_comparison}. Radiologists were asked to arrange the samples in order, from the most obvious to the most subtle nodules. Across cases, majority consensus ordering aligned with our subtlety scale in 80\% of cases, confirming a clear, clinically relevant progression in subtlety of generated nodules.

\begin{table}[h]
  \centering
  \scriptsize               
  \setlength{\tabcolsep}{4pt} 
  \renewcommand{\arraystretch}{0.85} 
  \caption{Radiologist evaluation of characteristic-specific LoRA modules.}
  \label{tab:characteristic-accuracy}
  \begin{tabular}{@{} l r @{}} 
    \toprule
    \textbf{Nodule characteristic} & \textbf{Agreement (\%)} \\
    \midrule
    Calcification          & 80 \\
    Regular border         & 90 \\
    Irregular border       & 100 \\
    Homogeneous texture    & 90 \\
    Inhomogeneous texture  & 100 \\
    \bottomrule
  \end{tabular}
\end{table}

\begin{figure}[t]
  \centering
  \setlength{\tabcolsep}{4pt}
  \renewcommand{\arraystretch}{1.0}

  \begin{tabular}{@{} c c c c @{}}
      & \scriptsize \textbf{(a)} 
      & \scriptsize \textbf{(b)} 
      & \scriptsize \textbf{(c)} \\[4pt]

      \rotatebox{90}{\textbf{ \scriptsize Linear Merge}} &
      \includegraphics[width=0.15\linewidth]{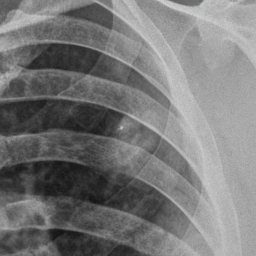} &
      \includegraphics[width=0.15\linewidth]{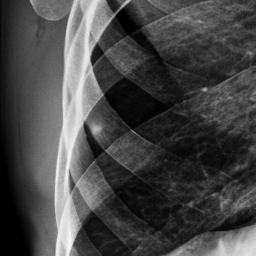} &
      \includegraphics[width=0.15\linewidth]{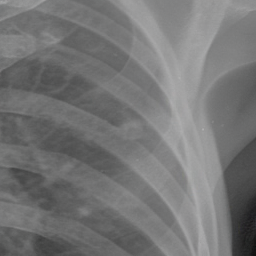} \\[6pt]

      \raisebox{0.8\height}{\rotatebox{90}{\scriptsize \textbf{Ours}}} &
      \includegraphics[width=0.15\linewidth]{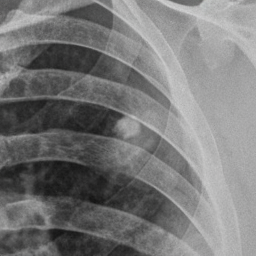} &
      \includegraphics[width=0.15\linewidth]{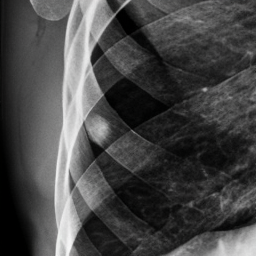} &
      \includegraphics[width=0.15\linewidth]{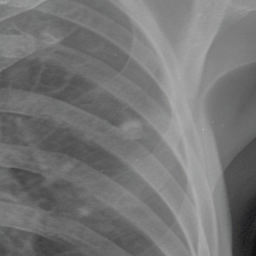} \\
  \end{tabular}

  \caption{Qualitative comparison between Linear Merge and our orthogonal merging across three nodule-attribute configurations: 
    (a) calcified + irregular, 
    (b) calcified + homogeneous, 
    (c) regular + homogeneous.  }
  \label{fig:lora-merge-rotated}
\end{figure}



\subsection{ Downstream Evaluation } 
\textbf{Diffusion Baseline Evaluation: } We evaluated the detection performance of models augmented with synthesized nodule from our baseline diffusion model. A Swin-Tiny\cite{liu2021swintransformerhierarchicalvision} encoder combined with a U-Net++\cite{zhou2018unetnestedunetarchitecture} architecture was trained jointly for nodule classification and segmentation. The training set consisted of approximately 10k real nodules with nearly 100k normal CXRs, supplemented with synthesized nodules. 

\noindent As shown in Table~\ref{tab:quantitative-eval-wide}, augmenting the training data with our synthetic nodules consistently improved both classification and segmentation performance across all test sets. These results highlight two important trends: (1) augmenting with synthetic nodules consistently boosts downstream detection performance across datasets, and (2) performance improvements plateau or slightly decline beyond an optimal level of augmentation, suggesting that carefully balanced integration of synthetic data maximizes its effectiveness.

\begin{table*}[h]  
  \centering
  \scriptsize
  \setlength{\tabcolsep}{4pt}
  \renewcommand{\arraystretch}{0.95}
  \caption{Quantitative evaluation of the effectiveness of different quantities of synthesized nodule data. Reported metrics include AUC and best IoU on three test sets.}
  \label{tab:quantitative-eval-wide}

  \begin{tabular}{@{} l cc cc cc @{}}
    \toprule
    \textbf{Train Data} 
      & \multicolumn{2}{c}{\textbf{In-house}} 
      & \multicolumn{2}{c}{\textbf{JSRT}} 
      & \multicolumn{2}{c}{\textbf{CheX-ray14}} \\
    
    & \textbf{AUC} & \textbf{IoU}
    & \textbf{AUC} & \textbf{IoU}
    & \textbf{AUC} & \textbf{IoU} \\
    \midrule

    10k real
      & 0.9705 & 0.3090
      & 0.8560 & 0.2475
      & 0.9008 & 0.5285 \\

    10k real + 2k syn
      & 0.9788 & 0.3222
      & 0.8639 & 0.2743
      & 0.9168 & 0.5293 \\

    10k real + 4k syn
      & 0.9780 & 0.3197
      & 0.8780 & 0.2589
      & 0.9245 & 0.5500 \\

    10k real + 6k syn
      & \textbf{0.9802} & \textbf{0.3247}
      & 0.8940 & 0.2894
      & 0.9315 & 0.5750 \\

    10k real + 8k syn
      & 0.9796 & 0.3274
      & 0.8864 & 0.2923
      & \textbf{0.9341} & \textbf{0.5954} \\

    10k real + 10k syn
      & 0.9801 & 0.3056
      & \textbf{0.9023} & \textbf{0.3091}
      & 0.9318 & 0.5613 \\

    \bottomrule
  \end{tabular}
\end{table*}

\noindent \textbf{Characteristic-Specific LoRA Adapters Evaluation: } We trained a Swin-Tiny~\cite{liu2021swintransformerhierarchicalvision} encoder with a multi-head classification module for all radiological characteristics, augmenting the training set with approximately 400 synthetic nodules per characteristic. The results in Table~\ref{tab:comp_5k} show that with augmentation the IoU score has improved across all characteristics on our in-house testset. We also evaluate our subtlety slider by showcasing improvements on JSRT subtlety dataset given in Appendix~\ref{sec:subtlety-eval}

\begin{table}[t]
\centering
\caption{Comparison of models against IOU scores trained with 5k real nodules versus 5k real nodules with 2k characteristic specific synthetic nodules across radiological features. 
}
\label{tab:comp_5k}
\setlength{\tabcolsep}{6pt}
\renewcommand{\arraystretch}{0.95}
\scriptsize
\begin{tabular}{@{}llccc@{}}
\toprule
 \textbf{Characteristic} & \textbf{5k Real} & \textbf{5k Real + 2k fake }\\
\midrule
\multirow{6}{*}{}
Nodule            & 0.2696 & \textbf{0.3002}\\
Calcification      & 0.2879 & \textbf{0.3199} \\
Regular Border    & 0.2941 & \textbf{0.3301} \\
Irregular Border  & 0.2733 & \textbf{0.3080} \\
Homogeneous        & 0.2695 & \textbf{0.3050} \\
Inhomogeneous      & 0.2706 & \textbf{0.2963} \\
\bottomrule
\end{tabular}
\end{table}


\subsection{Comparison with Existing Methods} 
To ensure a fair comparison, all baselines were trained on the same in-house dataset. We benchmark three families of generative approaches: GAN-based models, fill-based inpainting, and our Stage-2 diffusion framework (Figure~\ref{fig:your_label}). For inpainting, we include CR-Fill~\cite{zhao2021crfill}, the top performer in the NODE21 Generation Track~\cite{Sogancioglu2024NODE21}, given its strong CXR inpainting performance. For GANs, we evaluate ACGAN~\cite{odena2017acgan} and ReACGAN~\cite{lee2021reacgan}, widely used class-conditional frameworks. To assess the impact of synthetic nodules, we augmented the training data with 10k generated samples from each method and measured classification AUC on JSRT and ChestX-ray14 (Table~\ref{tab:combined-metrics2}). Although all augmentations improved over using 10k real samples alone, diffusion-based augmentation achieved the highest gains of 0.9023 AUC on JSRT and 0.9318 on ChestX-ray14, demonstrating its effectiveness for downstream detection.

\begin{table}[h]
\centering
\caption{Comparison of effect of synthetic-data augmentation on nodule AUC scores across \textit{ChestX-ray14} and \textit{JSRT}. }
\label{tab:10k_configs}
\setlength{\tabcolsep}{8pt}      
\renewcommand{\arraystretch}{1} 
\scriptsize
\begin{tabular}{lcc}
\toprule
\textbf{Augmentation} & \textbf{JSRT} & \textbf{ChestX-ray14} \\
\midrule
10k real                       & 0.8560 & 0.9008 \\
10k real + 10k ACGAN         & 0.8780 & 0.9281 \\
10k real + 10k ReACGAN       & 0.8808 & 0.9259 \\
10k real + 10k CRFILL       & 0.8786 & 0.9296 \\
10k real + 10k DiT-XL/2(Ours)& \textbf{0.9023} & \textbf{0.9318} \\
\bottomrule
\end{tabular}
\label{tab:combined-metrics2}
\end{table}

\section{Conclusion}
We introduced a novel diffusion-based framework for pulmonary nodule synthesis with characteristic-specific LoRA adapters, and an orthogonality constrained LoRA merging strategy. Experiments show that our method generates realistic and controllable nodules, outperforms GAN and inpainting-based baselines, and improves downstream CAD performance, with radiologist evaluations confirming clinical plausibility. Limitations include difficulty with some out-of-distribution generations by composition of LoRAs. Future work include taking merging to more than two characteristics.


\clearpage


\bibliography{main}

@String(ICCV= {Int. Conf. Comput. Vis.})

@String(AAAI = {AAAI})

@String(ICCV  = {ICCV})

@misc{lc_1,
author = {Bray, Freddie and Ferlay, Jacques and Soerjomataram, Isabelle and Siegel, Rebecca L. and Torre, Lindsey A. and Jemal, Ahmedin},
title = {Global cancer statistics 2018: GLOBOCAN estimates of incidence and mortality worldwide for 36 cancers in 185 countries},
journal = {CA: A Cancer Journal for Clinicians},
volume = {68},
number = {6},
pages = {394-424},
doi = {https://doi.org/10.3322/caac.21492},
url = {https://acsjournals.onlinelibrary.wiley.com/doi/abs/10.3322/caac.21492},
eprint = {https://acsjournals.onlinelibrary.wiley.com/doi/pdf/10.3322/caac.21492},
year = {2018}
}

@article{Rogers2010,
  author    = {Rogers, T. K. and Muir, K. and Baragwanath, P. and Evans, R. and Hodgson, J. and Kelly, M. and Morris, J. and Shepherd, S. and Smith, E. and Stephenson, T. and Welch, L. and Wilkinson, S.},
  title     = {Primary care radiography in the early diagnosis of lung cancer: randomised controlled trial},
  journal   = {The British Journal of General Practice},
  year      = {2010},
  volume    = {60},
  number    = {571},
  pages     = {e315--e322},
  doi       = {10.3399/bjgp10X514729},
  url       = {https://pmc.ncbi.nlm.nih.gov/articles/PMC2842172/}
}

@article{Sogancioglu2024NODE21,
  author    = {Ecem Sogancioglu and Bram van Ginneken and Finn Behrendt and Marcel Bengs and Alexander Schlaefer and Miron Radu and Di Xu and Ke Sheng and Fabien Scalzo and Eric Marcus and Samuele Papa and Jonas Teuwen and Ernst Th. Scholten and Steven Schalekamp and Nils Hendrix and Colin Jacobs and Ward Hendrix and Clara I. Sánchez and Keelin Murphy},
  title     = {Nodule Detection and Generation on Chest X-rays: NODE21 Challenge},
  journal   = {IEEE Transactions on Medical Imaging},
  year      = {2024},
  volume    = {43},
  number    = {8},
  pages     = {2839--2853},
  url       = {https://arxiv.org/abs/2401.02192}
}

@misc{prabhakar2024lorasoupsmergingloras,
      title={LoRA Soups: Merging LoRAs for Practical Skill Composition Tasks}, 
      author={Akshara Prabhakar and Yuanzhi Li and Karthik Narasimhan and Sham Kakade and Eran Malach and Samy Jelassi},
      year={2024},
      eprint={2410.13025},
      archivePrefix={arXiv},
      primaryClass={cs.CL},
      url={https://arxiv.org/abs/2410.13025}, 
}

@misc{kong2024loraswitchboostingefficiencydynamic,
      title={LoRA-Switch: Boosting the Efficiency of Dynamic LLM Adapters via System-Algorithm Co-design}, 
      author={Rui Kong and Qiyang Li and Xinyu Fang and Qingtian Feng and Qingfeng He and Yazhu Dong and Weijun Wang and Yuanchun Li and Linghe Kong and Yunxin Liu},
      year={2024},
      eprint={2405.17741},
      archivePrefix={arXiv},
      primaryClass={cs.AI},
      url={https://arxiv.org/abs/2405.17741}, 
}

@misc{zhou2018unetnestedunetarchitecture,
      title={UNet++: A Nested U-Net Architecture for Medical Image Segmentation}, 
      author={Zongwei Zhou and Md Mahfuzur Rahman Siddiquee and Nima Tajbakhsh and Jianming Liang},
      year={2018},
      eprint={1807.10165},
      archivePrefix={arXiv},
      primaryClass={cs.CV},
      url={https://arxiv.org/abs/1807.10165}, 
}

@misc{seer_csr_1975_2014,
  key       = {Howlader et al.},
  editor    = {Howlader, N. and Noone, A. M. and Krapcho, M. and Miller, D. and Bishop, K. and Kosary, C. L. and Yu, M. and Ruhl, J. and Tatalovich, Z. and Mariotto, A. and Lewis, D. R. and Chen, H. S. and Feuer, E. J. and Cronin, K. A.},
  title     = {SEER Cancer Statistics Review, 1975-2014},
  publisher = {National Cancer Institute},
  address   = {Bethesda, MD},
  year      = {2017},
  url       = {https://seer.cancer.gov/csr/1975_2014/},
  note      = {Based on November 2016 SEER data submission, posted to the SEER web site, April 2017}
}

@misc{liu2021swintransformerhierarchicalvision,
      title={Swin Transformer: Hierarchical Vision Transformer using Shifted Windows}, 
      author={Ze Liu and Yutong Lin and Yue Cao and Han Hu and Yixuan Wei and Zheng Zhang and Stephen Lin and Baining Guo},
      year={2021},
      eprint={2103.14030},
      archivePrefix={arXiv},
      primaryClass={cs.CV},
      url={https://arxiv.org/abs/2103.14030}, 
}

@article{Balagurunathan2019,
  author    = {Balagurunathan, Y. and Chen, A. and Clark, K. and et al.},
  title     = {Quantitative Imaging Features Improve Discrimination of Pulmonary Nodules},
  journal   = {Scientific Reports},
  year      = {2019},
  volume    = {9},
  pages     = {44562},
  doi       = {10.1038/s41598-019-44562-z}
}

@misc{lc_5,
author = {Irvin, Jeremy and Rajpurkar, Pranav and Ko, Michael and Yu, Yifan and Ciurea-Ilcus, Silviana and Chute, Chris and Marklund, Henrik and Haghgoo, Behzad and Ball, Robyn and Shpanskaya, Katie and Seekins, Jayne and Mong, David A. and Halabi, Safwan S. and Sandberg, Jesse K. and Jones, Ricky and Larson, David B. and Langlotz, Curtis P. and Patel, Bhavik N. and Lungren, Matthew P. and Ng, Andrew Y.},
title = {CheXpert: a large chest radiograph dataset with uncertainty labels and expert comparison},
year = {2019},
isbn = {978-1-57735-809-1},
publisher = {AAAI Press},
url = {https://doi.org/10.1609/aaai.v33i01.3301590},
doi = {10.1609/aaai.v33i01.3301590},
booktitle = {Proceedings of the Thirty-Third AAAI Conference on Artificial Intelligence and Thirty-First Innovative Applications of Artificial Intelligence Conference and Ninth AAAI Symposium on Educational Advances in Artificial Intelligence},
articleno = {73},
numpages = {8},
location = {Honolulu, Hawaii, USA},
series = {AAAI'19/IAAI'19/EAAI'19}
}

@article{lc_7,
  author    = {Khan, Ali Nawaz and Al-Jahdali, H. H. and Allen, C. M. and Irion, K. L. and Al Ghanem, S. and Koteyar, S. S.},
  title     = {The calcified lung nodule: What does it mean?},
  journal   = {Annals of Thoracic Medicine},
  volume    = {5},
  number    = {2},
  pages     = {67-79},
  year      = {2010},
  doi       = {10.4103/1817-1737.62469}
}

@article{lc_8,
  author    = {Zhang, R. and Tian, P. and Qiu, Z. and Liang, Y. and Li, W.},
  title     = {The growth feature and its diagnostic value for benign and malignant pulmonary nodules met in routine clinical practice},
  journal   = {Journal of Thoracic Disease},
  volume    = {12},
  number    = {5},
  pages     = {2019--2030},
  year      = {2020},
  doi       = {10.21037/jtd-19-3591},
  url       = {https://doi.org/10.21037/jtd-19-3591}
}

@inproceedings{odena2017acgan,
  title={Conditional image synthesis with auxiliary classifier gans},
  author={Odena, Augustus and Olah, Christopher and Shlens, Jonathon},
  booktitle={Proceedings of the 34th International Conference on Machine Learning},
  pages={2642--2651},
  year={2017},
  organization={PMLR}
}

@inproceedings{lee2021reacgan,
  title={ReACGAN: Training Auxiliary Classifier GANs with a Reversely Architected Discriminator},
  author={Lee, Younggeun and Ahn, Namhyuk and Kim, Daijin},
  booktitle={Advances in Neural Information Processing Systems},
  volume={34},
  pages={20049--20060},
  year={2021}
}

@inproceedings{zhao2021crfill,
  title={Large Scale Image Completion via Co-Modulated Generative Adversarial Networks},
  author={Zhao, Shengyu and Liu, Hongyu and Lin, Jiaya and Zhu, Zhe and Loy, Chen Change},
  booktitle={Proceedings of the IEEE/CVF International Conference on Computer Vision},
  pages={12838--12847},
  year={2021}
}

@article{schultheiss2021,
  author    = {Schultheiss, M. and Sogancioglu, E. and Setio, A. and Schwyzer, M. and Candemir, S.},
  title     = {The nodule generation challenge: synthetic nodules in chest radiographs},
  journal   = {Scientific Reports},
  year      = {2021},
  volume    = {11},
  number    = {1},
  pages     = {16165},
  doi       = {10.1038/s41598-021-94750-z},
  url       = {https://doi.org/10.1038/s41598-021-94750-z}
}

@article{hanaoka2024,
  author    = {Hanaoka, Shunsuke and Nomura, Yusuke and Yoshikawa, Takashi and Nakao, Takuya and Takenaga, Takashi and Matsuzaki, Hiroshi and Yamamichi, Nobutake and Abe, Osamu},
  title     = {Detection of pulmonary nodules in chest radiographs: novel cost function for effective network training with purely synthesized datasets},
  journal   = {International Journal of Computer Assisted Radiology and Surgery},
  year      = {2024},
  volume    = {19},
  number    = {10},
  pages     = {1991--2000},
  doi       = {10.1007/s11548-024-03227-7},
  pmid      = {39003437},
  pmcid     = {PMC11442563},
  month     = {Oct},
  note      = {Epub 2024 Jul 13}
}

@article{wang2022synthetic,
  author    = {Wang, Shuo and Qiao, Yifan and Qian, Wei and Yang, Ying and Shi, Jun},
  title     = {Generation of synthetic ground glass nodules using generative adversarial networks},
  journal   = {European Radiology Experimental},
  year      = {2022},
  volume    = {6},
  number    = {1},
  pages     = {32},
  doi       = {10.1186/s41747-022-00311-y},
  url       = {https://doi.org/10.1186/s41747-022-00311-y}
}

@article{albert2009solitary,
  author    = {Albert, Richard H. and Russell, John J.},
  title     = {Evaluation of the solitary pulmonary nodule},
  journal   = {American Family Physician},
  year      = {2009},
  volume    = {80},
  number    = {8},
  pages     = {827--831},
  month     = {Oct},
  issn      = {0002-838X},
  pmid      = {19835344}
}

@misc{gandikota-2023,
	author = {Gandikota, Rohit and Materzynska, Joanna and Zhou, Tingrui and Torralba, Antonio and Bau, David},
	month = {11},
	title = {{Concept Sliders: LORA adaptors for precise control in diffusion models}},
	year = {2023},
	url = {https://arxiv.org/abs/2311.12092},
}

@inproceedings{dhariwal2021diffusion,
  title     = {Diffusion Models Beat GANs on Image Synthesis},
  author    = {Dhariwal, Prafulla and Nichol, Alexander Quinn},
  booktitle = {Advances in Neural Information Processing Systems},
  year      = {2021},
  url       = {https://proceedings.neurips.cc/paper/2021/hash/49ad23d1ec9fa4bd8d77d02681df5cfa-Abstract.html}
}

@article{mou2023t2iadapter,
  title   = {T2I-Adapters: Learning Adapters to Dig out More Controllable Ability for Text-to-Image Diffusion Models},
  author  = {Mou, Chong and Zhang, Lvmin and Rao, Anyi and Agrawala, Maneesh},
  journal = {arXiv preprint arXiv:2302.08453},
  year    = {2023},
  url     = {https://arxiv.org/abs/2302.08453}
}

@article{hertz2022prompt,
  title   = {Prompt-to-Prompt Image Editing with Cross Attention Control},
  author  = {Hertz, Amir and Mokady, Ron and Tenenbaum, Jonathan and Aberman, Kfir and Cohen-Or, Daniel and Pritch, Yael},
  journal = {arXiv preprint arXiv:2208.01626},
  year    = {2022},
  url     = {https://arxiv.org/abs/2208.01626}
}

@article{podell2023sdxl,
  title   = {SDXL: Improving Latent Diffusion Models for High-Resolution Image Synthesis},
  author  = {Podell, David and others},
  journal = {arXiv preprint arXiv:2307.01952},
  year    = {2023},
  url     = {https://arxiv.org/abs/2307.01952}
}

@article{zhang2023controlnet,
  title   = {Adding Conditional Control to Text-to-Image Diffusion Models},
  author  = {Zhang, Lvmin and Rao, Anyi and Agrawala, Maneesh},
  journal = {arXiv preprint arXiv:2302.05543},
  year    = {2023},
  url     = {https://arxiv.org/abs/2302.05543}
}

@article{yang2022survey,
  title   = {A Survey on Generative Diffusion Models},
  author  = {Yang, Ling and Zhang, Zhilong and Hong, Sihyeon and Xu, Wenqi and Zhao, Zhongjie and Li, Bo},
  journal = {arXiv preprint arXiv:2209.02646},
  year    = {2022},
  url     = {https://arxiv.org/abs/2209.02646}
}

@inproceedings{hu2021lora,
  title     = {LoRA: Low-Rank Adaptation of Large Language Models},
  author    = {Hu, Edward J. and Shen, Yelong and Wallis, Phillip and others},
  booktitle = {International Conference on Learning Representations},
  year      = {2021},
  url       = {https://arxiv.org/abs/2106.09685}
}

@article{ruiz2023dreambooth,
  title   = {DreamBooth: Fine Tuning Text-to-Image Diffusion Models for Subject-Driven Generation},
  author  = {Ruiz, Nataniel and Li, Yuanzhen and Jampani, Varun and others},
  journal = {arXiv preprint arXiv:2208.12242},
  year    = {2023},
  url     = {https://arxiv.org/abs/2208.12242}
}

@article{lora-diffusion-2023,
  title   = {LORA for Stable Diffusion: Low-Rank Adaptation for Fast Personalization},
  author  = {Koh, Jing Yu and others},
  journal = {arXiv preprint arXiv:2311.13435},
  year    = {2023},
  url     = {https://arxiv.org/abs/2311.13435}
}

@article{lyu2023lowrank,
  title   = {Low-Rank Adaptation for Fast Text-to-Image Diffusion Fine-Tuning},
  author  = {Lyu, Xueyan and others},
  journal = {arXiv preprint arXiv:2312.06632},
  year    = {2023},
  url     = {https://arxiv.org/abs/2312.06632}
}

@article{kufel-2023,
	author = {Kufel, Jakub and Bielowka, Michal and Rojek, Marcin and Mitrega, Adam and Lewandowski, Piotr and Cebula, Maciej and Krawczyk, Dariusz and Bielowka, Marta and Kondol, Dominika and Bargiel-Laczek, Katarzyna and Paszkiewicz, Iga and Czogalik, Lukasz and Kaczynska, Dominika and Woclaw, Aleksandra and Gruszczynska, Katarzyna and Nawrat, Zbigniew},
	journal = {Journal of Personalized Medicine},
	month = {9},
	number = {10},
	pages = {1426},
	title = {{Multi-Label classification of chest x-ray abnormalities using transfer learning techniques}},
	volume = {13},
	year = {2023},
	doi = {10.3390/jpm13101426},
	url = {https://doi.org/10.3390/jpm13101426},
}

@article{shiraishi1996computer,
  title={Computer-aided diagnosis in medical imaging: operating characteristics and observer performance studies},
  author={Shiraishi, Junji and Luther, Gordon and Metz, Charles E and Doi, Kunio},
  journal={Journal of the Japanese Society of Radiological Technology},
  volume={52},
  number={9},
  pages={1184--1191},
  year={1996}
}

@misc{zhong2024multiloracompositionimagegeneration,
      title={Multi-LoRA Composition for Image Generation}, 
      author={Ming Zhong and Yelong Shen and Shuohang Wang and Yadong Lu and Yizhu Jiao and Siru Ouyang and Donghan Yu and Jiawei Han and Weizhu Chen},
      year={2024},
      eprint={2402.16843},
      archivePrefix={arXiv},
      primaryClass={cs.CV},
      url={https://arxiv.org/abs/2402.16843}, 
}

@misc{shah2023ziplorasubjectstyleeffectively,
      title={ZipLoRA: Any Subject in Any Style by Effectively Merging LoRAs}, 
      author={Viraj Shah and Nataniel Ruiz and Forrester Cole and Erika Lu and Svetlana Lazebnik and Yuanzhen Li and Varun Jampani},
      year={2023},
      eprint={2311.13600},
      archivePrefix={arXiv},
      primaryClass={cs.CV},
      url={https://arxiv.org/abs/2311.13600}, 
}

@misc{ouyang2025kloraunlockingtrainingfreefusion,
      title={K-LoRA: Unlocking Training-Free Fusion of Any Subject and Style LoRAs}, 
      author={Ziheng Ouyang and Zhen Li and Qibin Hou},
      year={2025},
      eprint={2502.18461},
      archivePrefix={arXiv},
      primaryClass={cs.CV},
      url={https://arxiv.org/abs/2502.18461}, 
}

@book{gundel-2021,
	author = {Gundel, Sebastian and Setio, Arnaud A. A. and Grbic, Sasa and Maier, Andreas and Comaniciu, Dorin},
	booktitle = {Informatik aktuell},
	publisher = {Springer},
	month = {1},
	pages = {288},
	title = {{Abstract: Extracting and Leveraging Nodule Features with Lung Inpainting for Local Feature Augmentation}},
	year = {2021},
	doi = {10.1007/978-3-658-33198-6\{_}68}

@article{shen-2022,
	author = {Shen, Zhenrong and Ouyang, Xi and Xiao, Bin and Cheng, Jie-Zhi and Shen, Dinggang and Wang, Qian},
	journal = {Medical Image Analysis},
	month = {12},
	pages = {102708},
	title = {{Image synthesis with disentangled attributes for chest X-ray nodule augmentation and detection}},
	volume = {84},
	year = {2022},
	doi = {10.1016/j.media.2022.102708},
	url = {https://doi.org/10.1016/j.media.2022.102708},
}

@misc{sogancioglu2018chestxrayinpaintingdeep,
      title={Chest X-ray Inpainting with Deep Generative Models}, 
      author={Ecem Sogancioglu and Shi Hu and Davide Belli and Bram van Ginneken},
      year={2018},
      eprint={1809.01471},
      archivePrefix={arXiv},
      primaryClass={cs.GR},
      url={https://arxiv.org/abs/1809.01471}, 
}

@article{litjens-2010,
	author = {Litjens, G. J. S. and Hogeweg, L. and Schilham, A. M. R. and De Jong, P. A. and Viergever, M. A. and Van Ginneken, B.},
	journal = {Lecture notes in computer science},
	month = {1},
	pages = {396--403},
	title = {{Simulation of nodules and diffuse infiltrates in chest radiographs using CT templates}},
	year = {2010},
	doi = {10.1007/978-3-642-15745-5\{_}49}

@article{behrendt2023systematic,
  title   = {A systematic approach to deep learning-based nodule detection in chest radiographs},
  author  = {Behrendt, F. and Bengs, M. and Bhattacharya, D. and others},
  journal = {Scientific Reports},
  volume  = {13},
  number  = {1},
  pages   = {10120},
  year    = {2023},
  publisher = {Nature Publishing Group},
  doi     = {10.1038/s41598-023-37270-2},
  url     = {https://doi.org/10.1038/s41598-023-37270-2}
}

@article{Nakayama1952Orthogonality,
  author    = {Tadasi Nakayama},
  title     = {Orthogonality Relation for Frobenius-and Quasi-Frobenius-Algebras},
  journal   = {Proceedings of the American Mathematical Society},
  year      = {1952},
  volume    = {3},
  pages     = {183--195},
}

@misc{zheng2025decoupleorthogonalizedatafreeframework,
      title={Decouple and Orthogonalize: A Data-Free Framework for LoRA Merging}, 
      author={Shenghe Zheng and Hongzhi Wang and Chenyu Huang and Xiaohui Wang and Tao Chen and Jiayuan Fan and Shuyue Hu and Peng Ye},
      year={2025},
      eprint={2505.15875},
      archivePrefix={arXiv},
      primaryClass={cs.CV},
      url={https://arxiv.org/abs/2505.15875}, 
}

@misc{ho2022classifierfreediffusionguidance,
      title={Classifier-Free Diffusion Guidance}, 
      author={Jonathan Ho and Tim Salimans},
      year={2022},
      eprint={2207.12598},
      archivePrefix={arXiv},
      primaryClass={cs.LG},
      url={https://arxiv.org/abs/2207.12598}, 
}

@article{peebles-2023,
	author = {Peebles, William and Xie, Saining},
	journal = {2021 IEEE/CVF International Conference on Computer Vision (ICCV)},
	month = {10},
	title = {{Scalable Diffusion Models with Transformers}},
	year = {2023},
	doi = {10.1109/iccv51070.2023.00387},
	url = {https://doi.org/10.1109/iccv51070.2023.00387},
}

@misc{zhang2025lorireducingcrosstaskinterference,
      title={LoRI: Reducing Cross-Task Interference in Multi-Task Low-Rank Adaptation}, 
      author={Juzheng Zhang and Jiacheng You and Ashwinee Panda and Tom Goldstein},
      year={2025},
      eprint={2504.07448},
      archivePrefix={arXiv},
      primaryClass={cs.LG},
      url={https://arxiv.org/abs/2504.07448}, 
}

@misc{gu2023mixofshowdecentralizedlowrankadaptation,
      title={Mix-of-Show: Decentralized Low-Rank Adaptation for Multi-Concept Customization of Diffusion Models}, 
      author={Yuchao Gu and Xintao Wang and Jay Zhangjie Wu and Yujun Shi and Yunpeng Chen and Zihan Fan and Wuyou Xiao and Rui Zhao and Shuning Chang and Weijia Wu and Yixiao Ge and Ying Shan and Mike Zheng Shou},
      year={2023},
      eprint={2305.18292},
      archivePrefix={arXiv},
      primaryClass={cs.CV},
      url={https://arxiv.org/abs/2305.18292}, 
}

\appendix
\section{Pseudocode}
\label{app:pseudocode}

\subsection{Training Pipeline for DiT backbone}


\begin{algorithm}[H]
\small
\caption{Training Pipeline for Mask-Conditioned DiT Backbone}
\label{alg:dit-training}
\begin{algorithmic}[1]
  \State \textbf{Input:} Dataset $\mathcal{D}=\{(x,m)\}$ where $x$ is a CXR patch and $m$ is a binary nodule mask.
  \State \textbf{Initialize:} VAE encoder--decoder, DiT backbone $f_\theta$, diffusion schedule $\{\alpha_t\}_{t=1}^T$, optimizer.
    \State \textbf{Repeat until convergence:}
    \State \hspace{0.5em} Sample $(x,m)\sim\mathcal{D}$.
    \State \hspace{0.5em}  $z_0 \leftarrow \mathrm{VAE.encode}(x)$.
    \State  \hspace{0.5em} $c \leftarrow \mathrm{process}(m)$.
    \State \hspace{0.5em}  Sample $t\sim\{1,\dots,T\}$ and $\;\epsilon\sim\mathcal{N}(0,I)$.
    \State  \hspace{0.5em} Form noisy latent:
    \Statex\hspace{\algorithmicindent} $z_t \leftarrow \sqrt{\alpha_t}\,z_0 + \sqrt{1-\alpha_t}\,\epsilon$.
    \State \hspace{0.5em}  Predict noise: $\hat{\epsilon}\leftarrow f_\theta(z_t,t,c)$.
    \State \hspace{0.5em}  Compute loss: $\mathcal{L}_{\mathrm{diff}} \leftarrow \| \epsilon - \hat{\epsilon} \|^2$.
    \State  \hspace{0.5em} Update parameters: $\theta \leftarrow \theta - \eta \nabla_\theta \mathcal{L}_{\mathrm{diff}}$.
  \State \textbf{Until convergence.}
  
\end{algorithmic}
\end{algorithm}

\subsection{Training Pipeline for LoRA Adapter}

\begin{algorithm}[h]
\small
\caption{LoRA Adapter Training}
\label{alg:lora-basic}
\begin{algorithmic}[1]
    \State \textbf{Freeze:} backbone parameters $\theta$; train LoRA parameters $\phi$ only
    \State \textbf{Repeat until convergence:}
    \State \hspace{1em} Sample $(x, m) \sim \mathcal{D}$
    \State \hspace{1em} $z_0 \gets \mathrm{VAE.encode}(x)$
    \State \hspace{1em} $c \gets \mathrm{process}(m)$
    \State \hspace{1em} Sample $t \sim \{1, \dots, T\}$ and $\epsilon \sim \mathcal{N}(0, I)$
    \State \hspace{1em} $z_t \gets \sqrt{\alpha_t}\, z_0 + \sqrt{1 - \alpha_t}\, \epsilon$
    \State \hspace{1em} $\hat{\epsilon} \gets f_{\theta,\phi}(z_t, t, c)$
    \State \hspace{1em} Compute $\mathcal{L}_{\mathrm{diff}}$ (plus optional $\mathcal{L}_{\mathrm{ortho}}$, $\mathcal{L}_{\mathrm{con}}$)
    \State \hspace{1em} Update LoRA parameters $\phi$ using the combined loss
\end{algorithmic}
\end{algorithm}

\subsection{Inference Algorithm (Conditional Sampling)}

\begin{algorithm}[H]
\small
\caption{Inference via Mask-Conditioned Reverse Diffusion}
\label{alg:dit-inference}
\begin{algorithmic}[1]
    \State \textbf{Input:} Binary mask $m$, trained backbone $f_\theta$, VAE decoder, diffusion schedule $\{\alpha_t, \sigma_t\}_{t=1}^T$
    \State Sample $z_T \sim \mathcal{N}(0, I)$
    \State $c \gets \mathrm{process}(m)$
    \State \textbf{For timesteps $t = T$ down to $1$:}
    \State \hspace{1em} $\hat{\epsilon} \gets f_\theta(z_t, t, c)$
    \State \hspace{1em} $\mu_\theta \gets \frac{1}{\sqrt{\alpha_t}}\bigl(z_t - (1 - \alpha_t)\hat{\epsilon}\bigr)$
    \State \hspace{1em} Sample $z_{t-1} \sim \mathcal{N}(\mu_\theta, \sigma_t^2 I)$
    \State $\hat{x} \gets \mathrm{VAE.decode}(z_0)$
    \State \textbf{Output:} Synthesized CXR patch $\hat{x}$
\end{algorithmic}
\end{algorithm}

\section{Experimental Setup}

\noindent \textbf{Base Diffusion Model  \&  Training Setup: }The DiT-XL/2 backbone contains 28 transformer blocks, each composed of attention and MLP components.  We initialize the DiT-XL/2 backbone from a publicly available pre-trained checkpoint. Full-resolution chest X-rays are standardized to $960 \times 960$ pixels, from which $256 \times 256$ nodule-centered patches are extracted. These patches are encoded into $32 \times 32$ latent representations using the StabilityAI VAE-FT-EMA. The diffusion process follows a 1000-step DDPM with a linear noise schedule, selected based on pilot experiments to ensure stable convergence and high-fidelity reconstructions. More details are given in the Appendix.

\noindent \textbf{LoRA Implementation: }
We build on the standard LoRA formulation described in Section 4.3. 
For all experiments, the adapter rank is fixed at $r=32$. 
The scaling factor is set to $\alpha=1.0$ for all characteristics, except for subtlety, where we adopt a variable scaling $\alpha(s) = 2^{2+s}$, with $s$ denoting the annotated subtlety levels (1--5). 
The down-projection matrix is initialized with Kaiming uniform initialization, while the up-projection matrix is initialized with zeros. 


\noindent \textbf{LoRA Integration with DiT Architecture :} LoRA adapters are inserted into the DiT-XL/2 attention mechanism, with Query–Key–Value (QKV) projections as the primary adaptation targets 
and output projections as secondary targets. 
This design allows efficient characteristic-specific adaptation while keeping the 675M-parameter backbone frozen. The total LoRA parameters per characteristic amount to approximately 6.2M parameters, representing only 0.9\% of the base model's parameters.


\subsection{Hyperparameters}

\begin{table}[h]
\centering
\caption{Backbone Training Hyperparameters (DiT-XL/2)}
\label{tab:backbone-hparams}
\begin{tabular}{l l}
\toprule
\textbf{Parameter} & \textbf{Value} \\
\midrule
Model type              & DiT-XL/2 \\
Input size (latent)     & 32 \\
Patch size              & 2 \\
Hidden size             & 1152 \\
Depth                   & 28 \\
Attention heads         & 16 \\
MLP ratio               & 4.0 \\
Epochs                  & 50{,}000 \\
Batch size              & 80 \\
Learning rate           & $1\times 10^{-4}$ \\
Optimizer               & AdamW ($\beta_1=0.9$, $\beta_2=0.999$, $\epsilon=10^{-8}$) \\
Noise schedule          & Linear, $T=1000$ steps \\
Context conditioning    & Concat-transformer with SpatialConv+Drop \\
CFG scale               & 4.0 \\
\bottomrule
\end{tabular}
\end{table}

\begin{table}[h]
\centering
\caption{LoRA Adapter Training Hyperparameters}
\label{tab:lora-hparams}
\begin{tabular}{l l}
\toprule
\textbf{Parameter} & \textbf{Value} \\
\midrule
Rank ($r$)            & 32 \\
Scaling $\alpha$      & 1.0 \\
Training method       & noxattn \\
Epochs                & 150 \\
Batch size            & 200 \\
Learning rate         & $5\times 10^{-5}$ \\
Optimizer             & AdamW \\
Scheduler             & Constant \\
Weight decay          & 0.01 \\
Precision             & FP32 \\
\bottomrule
\end{tabular}
\end{table}

\begin{table}[h]
\centering
\caption{Contrastive Fine-Tuning Hyperparameters}
\label{tab:contrastive-hparams}
\begin{tabular}{l l}
\toprule
\textbf{Parameter} & \textbf{Value} \\
\midrule
Temperature $\tau$       & 0.07 \\
Margin                   & 1.0 \\
Feature dimension        & 1152 \\
Pooling                  & Mean pooling \\
Feature normalization    & L2 norm \\
Positive pairs           & Subtlety-based grouping \\
Negative pairs           & Random shuffle \\
Min pos/neg samples      & 2 each per batch \\
\bottomrule
\end{tabular}
\end{table}

\begin{table}[h]
\centering
\caption{Orthogonality Regularization Hyperparameters}
\label{tab:ortho-hparams}

\scriptsize       
\setlength{\tabcolsep}{4pt}   
\renewcommand{\arraystretch}{0.9}  

\begin{tabular}{@{}l l@{}}
\toprule
\textbf{Parameter} & \textbf{Value} \\
\midrule
Ortho weight            & 0.5 \\
Loss type               & Frobenius norm \\
Target                  & Identity matrix \\
Adapter pairs           & (calcified, homogeneous), (irregular, homogeneous), (calcified, irregular) \\
Training strategy       & Alternating batches + joint optimization \\
Gradient accumulation   & 2 \\
\bottomrule
\end{tabular}
\end{table}



\begin{equation}
\mathcal{L}_{\text{inpaint}} = 
\mathbb{E}_{x,m,\epsilon,t} \left[
\left\| \epsilon -
\epsilon_\theta\left(
\sqrt{\bar{\alpha}_t}(x \odot (1-m)) +
\sqrt{1-\bar{\alpha}_t}\,\epsilon, m, t
\right) \right\|^2
\right]
\end{equation}

\begin{equation}
\mathcal{L}_{\text{contrastive}} = 
-\sum_i \log
\frac{\exp\left(\text{sim}(z_i, z_i^+)/\tau\right)}
{\sum_j \exp\left(\text{sim}(z_i, z_j^-)/\tau\right)}
\end{equation}

\begin{equation}
\mathcal{L}_{\text{ortho}} =
\left\| W_a^\top W_b - I \right\|_F^2
\end{equation}

\begin{table}[H]
\centering
\small 
\setlength{\tabcolsep}{3pt} 
\caption{LoRA Parameter Distribution in DiT-XL/2}
\begin{tabular}{lccc}
\toprule
\textbf{Component} & \textbf{Dimensions} & \textbf{Per Block} & \textbf{Total} \\
\midrule
QKV Adapters      & $32\times1152 + 3456\times32$ & 147,456 & 4,128,768 \\
Proj Adapters     & $32\times1152 + 1152\times32$ & 73,728  & 2,064,384 \\
\midrule
\textbf{Total per Adapter} & -- & 221,184 & \textbf{6,193,152} \\
\bottomrule
\end{tabular}
\label{tab:lora_params}
\end{table}

\section{Characteristic Definitions}

\textbf{Homogeneity:} Homogeneity refers to the uniformity of radiographic density (intensity levels) within a pulmonary nodule throughout its entire cross-sectional area. In contrast, non-homogeneous (heterogeneous) nodules show uneven density patterns, with some areas appearing brighter and others darker, often indicative of malignancy. 

\noindent \textbf{Boundary morphology (regular vs. irregular):} Regular nodules have smooth, well-defined borders with clear demarcation from lung tissue. Irregular nodules show variable characteristics, including spiculated edges, lobulated contours, or poorly defined margins that blend with surrounding tissue—commonly associated with malignancy. 

\noindent \textbf{Calcification:} Calcified nodules are characterized by high radiographic intensity and are generally smaller in size. Calcification, resulting from calcium deposits, is often associated with benign nodules and appears brighter than the surrounding tissue. 

\noindent \textbf{Subtlety:} Subtle nodules refer to pulmonary lesions that demonstrate minimal radiographic contrast with surrounding lung parenchyma, making them challenging to detect on standard chest X-ray imaging. These nodules typically exhibit low-density characteristics with opacity levels that closely approximate normal lung tissue, resulting in poor visual conspicuity against the background. From the subtlety distribution analysis of our annotated scores, nodules cover a wide spectrum, with most having low subtlety scores (more subtle) and fewer having high scores (more visible).

\noindent \textbf{Nodule size:} Nodule size represents a critical malignancy risk factor, with larger nodules generally indicating higher malignancy probability. However, characteristics typically manifest in combination rather than isolation. Benign nodules commonly present as homogeneous lesions with regular margins and calcification, while malignant nodules frequently exhibit heterogeneous texture with ill-defined borders. The inherent difficulty of detecting subtle nodules underscores the importance of synthetic data generation that incorporates multiple co-occurring characteristics for improved detection and diagnosis models.
\section{Radiologist Evaluation Protocol}

\subsection*{Task 1: Real vs.\ Synthetic Nodule Assessment}
\textbf{Background:}  
To assess the visual realism of synthetic nodules, we inserted AI-generated nodules into authentic chest X-rays and asked radiologists to distinguish them from real clinical nodules.  

\textbf{Data:}  
The evaluation set comprised \textbf{50 chest X-ray images} containing nodules:
\begin{itemize}
    \item \textbf{Real nodules:} Pathological findings from patient scans.  
    \item \textbf{Synthetic nodules:} AI-generated nodules blended into authentic radiographs.  
\end{itemize}

\textbf{Procedure:}  
Radiologists reviewed each image and gave a binary response:
\begin{itemize}
    \item \textbf{Yes (Real):} Nodule appears clinically genuine.  
    \item \textbf{No (Synthetic):} Nodule appears AI-generated.  
\end{itemize}

\textbf{Goal:}  
This task measured how convincing AI-generated nodules appear relative to real clinical nodules.  

\subsection*{Task 2: Characteristic Verification}
\textbf{Background:}  
We next evaluated whether synthetic nodules accurately reflected specific radiological characteristics.  

\textbf{Data:}  
Five morphological categories were tested, with \textbf{10 images per characteristic}:  
Calcified, Homogeneous, Inhomogeneous, Irregular Border, and Regular Border.  
Each set of images was organized into a separate folder with an annotation sheet.  

\textbf{Procedure:}  
For each image, radiologists judged whether the nodule matched the stated feature:
\begin{itemize}
    \item \textbf{Yes:} Exhibits the described characteristic.  
    \item \textbf{No:} Does not match the characteristic.  
\end{itemize}

\textbf{Goal:}  
This task evaluated the morphological fidelity of AI-generated nodules across clinically relevant categories.  

\subsection*{Task 3: Subtlety Ranking}
\textbf{Background:}  
Subtlety, or how easily a nodule can be perceived, is clinically important. We generated nodules at different subtlety levels using our diffusion-based framework.  

\textbf{Data:}  
Radiologists received \textbf{20 sets of images}, each containing \textbf{3 versions of the same nodule} rendered at increasing levels of subtlety.  

\textbf{Procedure:}  
Within each set, radiologists ranked the three images:
\begin{itemize}
    \item Lowest Subtlety (hardest to detect) $\rightarrow$ Highest Subtlety (easiest to detect).  
\end{itemize}

\textbf{Goal:}  
This task tested whether the generative model produced nodules with perceptible and clinically meaningful differences in subtlety.  

\subsection*{Summary}
Together, these tasks: (1) Real vs.\ Synthetic classification, (2) Characteristic verification, and (3) Subtlety ranking, provided a comprehensive evaluation of realism, morphological fidelity, and perceptual detectability. This structured protocol ensured rigorous clinical validation of AI-generated nodules.

\section{Results And Analysis}
\subsection{Subtlety LoRA Evaluation }
\label{sec:subtlety-eval}
We generated subtle nodules  with Subtlety LoRA($\alpha<24$) and assessed their impact on classification performance using the JSRT dataset, which provides a 5-level subtlety grading. As shown in Table~\ref{tab:subtlety-reversed}, at the highest subtlety level (S1), sensitivity increased by 12\%  at Youden index threshold.
\begin{table}[h]
\centering
\scriptsize
\setlength{\tabcolsep}{3pt}
\renewcommand{\arraystretch}{0.75}
\caption{JSRT classification accuracy across subtlety levels (S1 = most subtle , S5 = least subtle ).}
\label{tab:subtlety-reversed}
\begin{tabular}{l c c c c c}
\toprule
\textbf{Train} & \textbf{S5} & \textbf{S4} & \textbf{S3} & \textbf{S2} & \textbf{S1} \\
\midrule
10k Real            & 100\% & 96\% & 70\% & 69\% & 40\% \\
10k Real + 12k Fake & 100\% & 100\% & 76\% & 76\% & 52\% \\
\bottomrule
\end{tabular}
\end{table}

\subsection{Comparison with Existing Methods}
Table~\ref{tab:combined-metrics} reports quantitative scores for both full-patch synthesis and masked-patch inpainting. Across all metrics, our method consistently outperforms the GAN and fill-based methods, demonstrating its superiority in synthesizing realistic lung nodules.
\begin{table}[h]
\scriptsize
\setlength{\tabcolsep}{3pt}
\renewcommand{\arraystretch}{0.8}
\centering
\caption{Comparison of generation methods on \textit{ChestX-ray14}}
\begin{tabular}{l ccc ccc}
\toprule
\multirow{2}{*}{\textbf{Method}} &
\multicolumn{3}{c}{\textbf{Full}} &
\multicolumn{3}{c}{\textbf{Masked}} \\
\cmidrule(lr){2-4} \cmidrule(lr){5-7}
 & \textbf{PSNR} & \textbf{SSIM} & \textbf{FID} & \textbf{PSNR} & \textbf{SSIM} & \textbf{FID} \\
\midrule
ACGAN      & 37.18 & 0.916 & 0.534 & 26.51 & 0.768 & 0.831 \\
ReACGAN    & 37.44 & 0.916 & 0.604 & 27.54 & 0.786 & 1.227 \\
CR-Fill    & 37.93 & 0.918 & 0.522 & 29.50 & 0.830 & 0.781 \\
DiT-XL/2 (Ours) & \textbf{38.74} & \textbf{0.920} & \textbf{0.390} & \textbf{34.26} & \textbf{0.892} & \textbf{0.475} \\
\bottomrule
\end{tabular}

\label{tab:combined-metrics}
\end{table}

\subsection{\textcolor{blue}{Comparison of CFG Control versus Separate LoRA for Label Guidance}}
\label{sec:cfgvslora}
Using classifier-free guidance (CFG) to control both nodule characteristics and the mask leads to suboptimal adherence to characteristic-specific attributes. To evaluate this, we perform an ablation comparing models trained with CFG-based label control against our approach using separate LoRA adapters. We compute FID scores for each characteristic, and as shown in Table~\ref{tab:cfg-vs-lora}, LoRA-based control yields consistently lower FID values, indicating stronger characteristic fidelity.

\begin{table}[h]
\centering
\caption{FID comparison between CFG-based label control and LoRA-based control across nodule characteristics.}
\label{tab:cfg-vs-lora}
\scriptsize
\setlength{\tabcolsep}{6pt}
\renewcommand{\arraystretch}{0.9}

\begin{tabular}{@{}lcc@{}}
\toprule
\textbf{Characteristic} & \textbf{CFG Control (FID)} & \textbf{LoRA Control (FID)} \\
\midrule
Calcification         & 15.87 $\pm$ 1.51 & 2.29 $\pm$ 0.09 \\
Regular Border        & 4.04  $\pm$ 0.84 & 1.96  $\pm$ 0.12 \\
Irregular Border      & 4.19  $\pm$ 0.59 & 2.90 $\pm$ 0.03 \\
Homogeneous Texture   & 8.48  $\pm$ 1.50 & 5.71 $\pm$ 0.36 \\
Inhomogeneous Texture & 12.58 $\pm$ 1.94 & 8.13 $\pm$ 0.28 \\
\bottomrule
\end{tabular}
\end{table}

\subsection{\textcolor{blue}{Radiologist evaluation with confidence intervals}}
\paragraph{Pooled vs.\ majority-vote metrics.}
We report results using two aggregation schemes across radiologists.
\emph{Pooled (vote-level)} metrics treat each radiologist response as one independent vote. If there are $N$ cases (images or ranking sets) and $R$ radiologists, the pooled denominator is $n = N \times R$. The pooled rate answers: ``Across all individual ratings, how often was the target response selected?''
\emph{Majority-vote (case-level)} metrics first aggregate the $R$ votes per case into a single panel decision (e.g., $\geq 2/3$ agreement), and then compute performance across cases with denominator $n=N$. The majority-vote rate answers: ``For how many cases did the panel agree with the target outcome?'' This provides an image level measure of consensus. We also provide 95\% confidence intervals ( Wilson Score )

\paragraph{Task 1: Real vs.\ synthetic}
Radiologists were presented with real and synthetic samples and asked to judge whether each sample appears \emph{real} or \emph{synthetic}. We report (i) the fraction of real images judged as real and (ii) the fraction of synthetic images judged as real, using both pooled and majority-vote aggregation. The results are  shown in  table \ref{task1}.

\paragraph{Task 2: Characteristic verification}
Radiologists were shown generated nodules targeting a specific radiological characteristic (e.g., border type, homogeneity, calcification) and asked whether the target characteristic is present. We report pooled and majority-vote ``Yes'' rates per characteristic to quantify how reliably the intended attribute is recognized by experts. The results are  shown in table \ref{task2}. 

\paragraph{Task 3: Subtlety ordering}
Radiologists were presented with small sets generated at different subtlety levels and asked to rank/order them by subtlety. Each set is scored as \emph{correct} vs.\ \emph{incorrect} ordering, and we report pooled and majority vote ranking accuracy, measuring how consistently the intended subtlety control aligns with expert perception. The results are  shown in table \ref{task3}. 

\begin{table}[H]
\centering
\caption{Task 1: Real vs Synthetic summary with 95\% CI}
\renewcommand{\arraystretch}{1.3} 
\begin{tabular}{@{}p{5.5cm}cccc@{}}
\toprule
\multirow{2}{*}{\textbf{Metric}} & \multicolumn{2}{c}{\textbf{Real images}} & \multicolumn{2}{c}{\textbf{Synthetic images}} \\ 
\cmidrule(lr){2-3} \cmidrule(l){4-5} 
 & \textbf{Rate} & \textbf{95\% CI} & \textbf{Rate} & \textbf{95\% CI} \\ 
\midrule
Pooled ``Looks real'' rate & 
0.867 & $[0.703, 0.947]$ & 
0.700 & $[0.521, 0.833]$ \\ 
\addlinespace
Majority-vote ``Looks real'' rate & 
0.900 & $[0.596, 0.982]$ & 
0.800 & $[0.490, 0.943]$ \\ 
\bottomrule
\label{task1}
\end{tabular}
\end{table}

\begin{table}[ht]
\centering
\caption{Task 2: Characteristic verification Pooled + Majority-vote with 95\% CI}
\renewcommand{\arraystretch}{1.3}
\begin{tabular}{@{}lcccc@{}}
\toprule
\multirow{2}{*}{\textbf{Characteristic}} & \multicolumn{2}{c}{\textbf{Pooled ``Yes'' rate}} & \multicolumn{2}{c}{\textbf{Majority-vote ``Yes'' rate}} \\ 
\cmidrule(lr){2-3} \cmidrule(l){4-5} 
 & \textbf{Rate (k/n)} & \textbf{95\% CI} & \textbf{Rate (k/n)} & \textbf{95\% CI} \\ 
\midrule
Homogeneous & $24/30 = 0.800$ & $[0.627, 0.905]$ & $9/10 = 0.900$ & $[0.596, 0.982]$ \\
Inhomogeneous & $22/27 = 0.815$ & $[0.633, 0.918]$ & $9/9 = 1.000$ & $[0.701, 1.000]$ \\
Irregular border & $27/30 = 0.900$ & $[0.744, 0.965]$ & $10/10 = 1.000$ & $[0.722, 1.000]$ \\
Regular border & $28/30 = 0.933$ & $[0.787, 0.982]$ & $9/10 = 0.900$ & $[0.596, 0.982]$ \\
Calcified & $20/30 = 0.66$ & $[0.488, 0.808]$ & $8/10 = 0.800$ & $[0.490, 0.943]$ \\ 
\bottomrule
\label{task2}
\end{tabular}
\end{table}

\begin{table}[ht]
\centering
\caption{Task 3 (Subtlety ranking): Accuracy with 95\% CI}
\renewcommand{\arraystretch}{1.3}
\begin{tabular}{@{}p{6cm}ccc@{}}
\toprule
\textbf{Metric} & \textbf{Result (k/n)} & \textbf{Accuracy} & \textbf{95\% CI} \\ 
\midrule
Pooled & 
$20/30$ & $0.667$ & $[0.488, 0.808]$ \\ 
\addlinespace
Majority-vote & 
$8/10$ & $0.800$ & $[0.490, 0.943]$ \\ 
\bottomrule
\label{task3}
\end{tabular}
\end{table}
\subsection{\textcolor{blue}{Orthogonality Verification via Adapter Weight Analysis}}
\label{app:orthogonality_verification}

To verify that the proposed orthogonality loss increases subspace orthogonality rather than trivially shrinking adapter weights, we analyze the behavior of adapter weight matrices under different training configurations. In particular, we compare adapters trained separately against configurations where two adapters are trained jointly with Frobenius-norm regularization.

\noindent Table~\ref{tab:orthogonality_frobenius} reports the aggregate magnitude of the adapter weight matrices across multiple characteristic pairings. Despite the presence of Frobenius-norm regularization, the jointly trained adapters consistently exhibit higher weight magnitudes compared to separately trained counterparts.

\begin{table}[h]
\centering
\caption{Comparison of adapter weight magnitudes for separately trained adapters versus two adapters trained jointly with Frobenius-norm regularization.}
\label{tab:orthogonality_frobenius}
\begin{tabular}{lccc}
\hline
\textbf{Comparison} & \textbf{Separate} & \textbf{Two Adapters (Frobenius)} & \textbf{\% Change} \\
\hline
Calcified vs.\ Homogeneous & 329.01 & 430.89 & +30.97\% \\
Homogeneous vs.\ Calcified & 388.23 & 502.42 & +29.41\% \\
Irregular vs.\ Homogeneous & 539.28 & 711.84 & +32.00\% \\
Homogeneous vs.\ Irregular & 388.23 & 803.59 & +106.99\% \\
\hline
\end{tabular}
\end{table}

\noindent Notably, this increase in weight magnitude occurs even though earlier results show a reduction in the Frobenius norm of individual adapters. This indicates that the orthogonality constraint does not simply suppress adapter activations. Instead, it encourages adapters to occupy more distinct directions in parameter space, leading to improved subspace separation.

\noindent These findings support the claim that the proposed orthogonality loss promotes genuine representational disentanglement between adapters, rather than acting as a magnitude-reducing regularizer.

\subsection{\textcolor{blue}{Computational Cost Analysis}}
\label{app:computational_cost}

This section summarizes the computational requirements of the proposed method, including training time, inference time, and hardware usage.

\noindent All experiments were conducted on NVIDIA L40 GPUs (48\,GB memory). Backbone training was performed in two stages. In Stage~1, the DiT backbone was trained for approximately one week using four L40 GPUs. In Stage~2, training was continued for approximately two additional days on four L40 GPUs. After Stage~2, the backbone parameters were frozen for all subsequent experiments.

\noindent The DiT-XL/2 backbone contains approximately 682M parameters. During adapter training, we optimized only the LoRA parameters, with each adapter comprising 6.2M parameters (approximately 0.91\% of the backbone).Each LoRA adapter was trained for 150 epochs with a batch size of 25 on $256 \times 256$ nodule-centered patches. Training a single adapter required approximately 10 hours on four L40 GPUs. Jointly training two adapters with the proposed orthogonality constraint required approximately 20-30 hours on four L40 GPUs, depending on the characteristic pairing and dataset size.

\noindent Inference was performed with 250 DDIM sampling steps, with an average runtime of approximately 10 seconds per sample. Since the backbone remains frozen at inference time, the computational cost is dominated by diffusion sampling rather than adapter-specific operations.

\noindent  Overall, the proposed approach remains computationally practical, providing parameter-efficient adaptation and manageable inference cost relative to full backbone fine-tuning.

\subsection{\textcolor{blue}{Extending Orthogonality Loss beyond two characteristics}} 
The orthogonality-based merging objective is not restricted to two attributes—it generalizes directly to K adapters by enforcing orthogonality across all selected LoRA updates (e.g., pairwise across the set), so in principle the same framework can synthesize nodules with three or more characteristics simultaneously.
\begin{equation}
\mathcal{L}_{\mathrm{orth}}=\sum_{i\ne j}\lVert W_i^{\top}W_j\rVert_F^2
\end{equation}

We did not include more than 2 characteristic compositions or downstream augmentation experiments in this submission for two practical reasons. First, our available datasets do not contain sufficiently dense co-occurrence annotations (i.e., reliable labels for multiple attributes on the same nodule) to form training/evaluation splits that would support a fair quantitative study of higher-order compositions. Second, as the number of characteristics increases, many combinations become rare or clinically incompatible in practice, leading to very small sample sizes and unstable estimates for both synthesis evaluation and downstream detection benchmarking.

\end{document}